\documentclass[letterpaper, 10 pt, conference]{ieeeconf}  

\usepackage{amsmath,amsfonts}
\usepackage{algorithmic}
\usepackage{algorithm}
\usepackage{array}
\usepackage[caption=false,font=small,labelfont=sf,textfont=sf]{subfig}
\usepackage{textcomp}
\usepackage{stfloats}
\usepackage{url}
\usepackage{verbatim}
\usepackage{graphicx}
\usepackage{cite}
\usepackage[OT1]{fontenc}
\usepackage[colorlinks, linkcolor=blue, anchorcolor=blue, citecolor=blue]{hyperref}
\usepackage{amssymb}
\usepackage{multirow}
\usepackage{marvosym}

\graphicspath{{Figures/}}

\IEEEoverridecommandlockouts                              

\title{\LARGE \bf IDAGC: Adaptive Generalized Human-Robot Collaboration via Human Intent Estimation and Multimodal Policy Learning}

\author{Haotian Liu$^{1,*}$, Yuchuang Tong$^{1,*,}\textsuperscript{\Letter}$, Guanchen Liu$^{1}$, Zhaojie Ju$^{2}$, Zhengtao Zhang$^{1,3,}\textsuperscript{\Letter}$
\thanks{This work was supported by the National Natural Science Foundation of China (62303457, U21A20482), Beijing Municipal Natural Science Foundation, China (4252053), and Zhongke Huiling Robot Technology Co., LTD. (E5D11703).}
\thanks{$^{1}$ CAS Engineering Laboratory for Intelligent Industrial Vision, Institute of Automation, Chinese Academy of Sciences, Beijing, China, and also with the School of Artificial Intelligence, University of Chinese Academy of Sciences, Beijing, China. {\tt\small liuhaotian2021, yuchuang.tong, liuguanchen2023, zhengtao.zhang@ia.ac.cn}}
\thanks{$^{2}$ Key Laboratory for Biomedical Engineering of Ministry of Education, Zhejiang University, China, and College of Biomedical Engineering and Instrument Science, Zhejiang University, China. {\tt\small juzhaojie@icloud.com}}
\thanks{$^{3}$ Beijing Zhongke Huiling Robot Technology Co., LTD, Beijing, China.}
\thanks{$^*$ Contributing equally, $\textsuperscript{\Letter}$ Corresponding author.}}

\begin{document}

\maketitle
\thispagestyle{empty}
\pagestyle{empty}

\begin{abstract}
    In Human-Robot Collaboration (HRC), which encompasses physical interaction and remote cooperation, accurate estimation of human intentions and seamless switching of collaboration modes to adjust robot behavior remain paramount challenges. To address these issues, we propose an Intent-Driven Adaptive Generalized Collaboration (IDAGC) framework that leverages multimodal data and human intent estimation to facilitate adaptive policy learning across multi-tasks in diverse scenarios, thereby facilitating autonomous inference of collaboration modes and dynamic adjustment of robotic actions. This framework overcomes the limitations of existing HRC methods, which are typically restricted to a single collaboration mode and lack the capacity to identify and transition between diverse states. Central to our framework is a predictive model that captures the interdependencies among vision, language, force, and robot state data to accurately recognize human intentions with a Conditional Variational Autoencoder (CVAE) and automatically switch collaboration modes. By employing dedicated encoders for each modality and integrating extracted features through a Transformer decoder, the framework efficiently learns multi-task policies, while force data optimizes compliance control and intent estimation accuracy during physical interactions. Experiments highlights our framework's practical potential to advance the comprehensive development of HRC.

\end{abstract}

\section{Introduction}

Human-Robot Collaboration (HRC) plays a critical role in manufacturing, healthcare, and services \cite{xingImpedanceLearningHumanGuided2023, 10759561, yuAdaptiveConstrainedImpedanceControl2022}, necessitating that robots seamlessly collaborate by accurately estimating human intentions and dynamically adapting to evolving tasks and environments, thereby mitigating the cognitive and physical burdens of human operators.

\begin{figure}[!t] 
    \centering
    \includegraphics[width=0.485\textwidth]{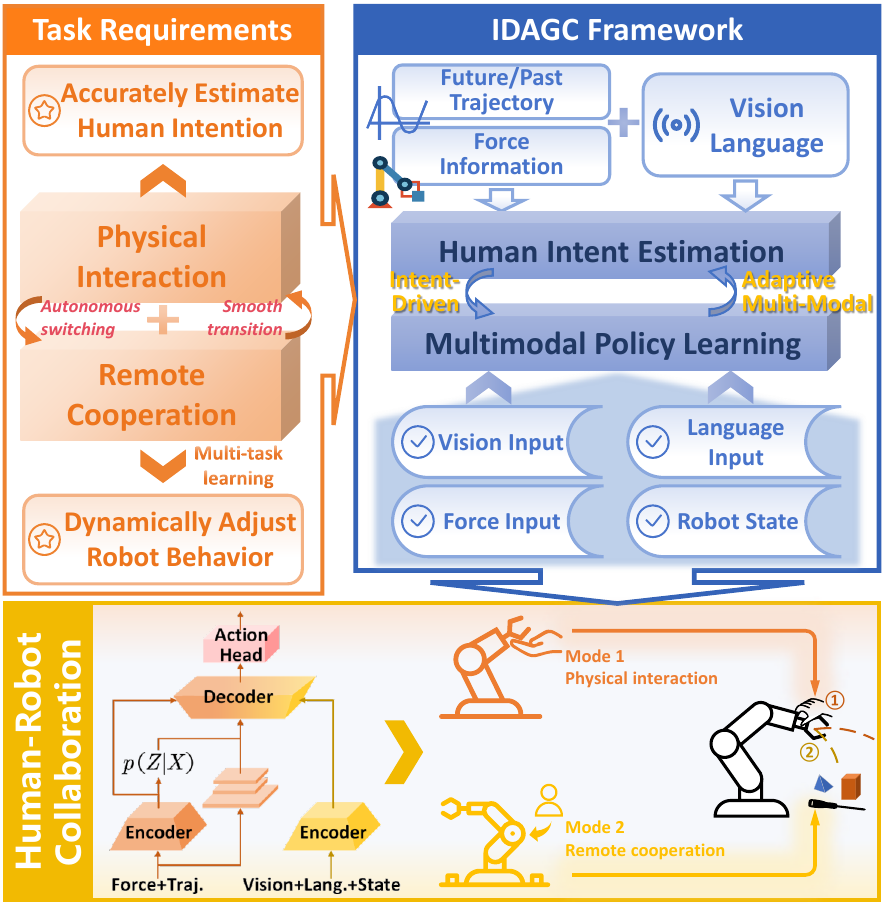}
    \caption{Overview of the IDAGC framework.}
    \label{fig:int}
\end{figure}

Contemporary HRC comprises physical Human-Robot Interaction (pHRI) and remote cooperation, as shown in Fig. \ref{fig:int}. However, existing methods rely on predefined modes and demonstrate limited capability in seamlessly transitioning based on task requirements and human intentions \cite{raviProactiveSafeHumanRobot2024}.
Moreover, they typically process single/double modalities among vision, language, and force, hindering its ability to capture the coupling characteristics of policies across multimodal information. Consequently, there is an urgent need for an adaptive framework that integrates remote cooperation with physical interaction through human-like intuition.

In pHRI, accurate prediction of future trajectories based on human intention is essential for efficient robot assistance and safety assurance \cite{mengHierarchicalHumanMotion2024}. In complex environments with potential hazards, such as unforeseen obstacles, sudden variations in human intent pose significant challenges for reliable estimation. Current methods predominantly rely on short-term motion data, like position and velocity, which constrains the ability to detect abrupt changes in intention and compromises prediction accuracy and safety \cite{liuFollowForceHaptic2023, zhangOnlineHumanDynamic2023}. Moreover, this overreliance undermines the effectiveness of intent estimation in long-term collaborations. Therefore, exploring long-term prediction methods that integrate human-applied forces is essential for enhancing accurate intent estimation.

In remote cooperation, robots must accurately respond to human guidance while autonomously executing multi-tasks, necessitating the integration of multimodal data, including vision, language, and robot state information. Furthermore, equipping robots with a broad repertoire of skills requires extensive experiential exposure and the capacity to learn from diverse data \cite{bharadhwajRoboAgentGeneralizationEfficiency2024}. A prevalent policy for multi-task training involves employing supervised learning on large datasets to synthesize universal policies \cite{kim2024openvla, zhangEXTRACTEfficientPolicy2024}. However, the interactive nature of these tasks renders data acquisition time and cost intensive, as it requires physical interaction with the environment \cite{hansenMoDemAcceleratingVisual2023}. Consequently, developing multi-task policies that can be effectively learned from a handful of demonstrations remains a significant challenge.

To address these challenges, we propose an Intent-Driven Adaptive Generalized Collaboration (IDAGC) framework that leverages multimodal data with human intent estimation to facilitate adaptive policy learning across diverse tasks and scenarios, thereby addressing the challenges of accurately estimating human intention and seamlessly integrating multimodal information in HRC. IDAGC overcomes the limitations of existing HRC methods \cite{bharadhwajRoboAgentGeneralizationEfficiency2024, NEURIPS2024_ff887781, cremerModelFreeOnlineNeuroadaptive2020, maHumanRobotCollaboration2024, franceschiLearningHumanMotion2023} by autonomously inferring collaboration modes and dynamically adjusting robot behaviors, thereby accommodating a broad spectrum of HRC scenarios, including pHRI and remote cooperation. Moreover, it effectively learns multi-task policies from a handful of demonstrations with enhanced accuracy and adaptability. Furthermore, the integration of force data refines accurate intent estimation and optimizes compliance during interactions, improving collaborative efficiency and safety. The main contributions of this paper are as follows.

\begin{itemize}
    \item [1)] IDAGC integrates human intent estimation with multimodal policy learning to autonomously infer collaboration modes and adjust robot behaviors, thereby enhancing HRC in complex and hazard-prone environments.
    
    \item [2)] The human intent estimation in IDAGC concurrently processes motion, force, and vision data, significantly improving prediction accuracy and equipping the robot with an understanding of human intentions.
    
    \item [3)] IDAGC facilitates multi-task policy learning with improved compliance and accuracy through dedicated encoders for different multimodal information within a unified architecture.

\end{itemize}

\section{Related Work}

{\noindent \bf Human Intent Estimation.} PHRI requires that robots accurately understand and respond to human needs through physical coupling\cite{shao2024constraintaware}. Current pHRI research focuses on recognizing and predicting human intentions using machine learning methods and on ensuring compliant collaboration via impedance/admittance control\cite{11049934, wangRoleDynamicAssignment2024}. Traditional intent estimation relies on high-precision force sensors to capture external human-applied forces, which indicate the direction of human motion\cite{Huang2005SharedNC}, while alternative methods decode neural signals such as electromyography or electroencephalography\cite{6943678}. Recent neural network-based methods, including a two-layer neural network by Cremer et al.\cite{cremerModelFreeOnlineNeuroadaptive2020} and a Bayesian Neural Network by Ma et al.\cite{maHumanRobotCollaboration2024}, have been applied for intent estimation. However, these short-term prediction techniques tend to underutilize prior information and are limited to next-step predictions, resulting in suboptimal local decisions. In contrast, long-term prediction methods such as Recurrent Neural Networks (RNNs) and Transformers have garnered attention due to their ability to process sequential data by incorporating contextual information from previous motion states, thereby enhancing estimation accuracy\cite{gaoHybridRecurrentNeural2023,songRobotTrajectronTrajectory2024, franceschiLearningHumanMotion2023}. Nevertheless, RNNs encounter challenges in handling long-term dependencies and in parallel processing, which can diminish prediction efficiency and accuracy.

Moreover, existing long-term prediction methods primarily focus on motion data, which limits their ability to detect shifts in human intentions, restricting the robot’s ability to adjust its behavior accordingly\cite{liu2025dtrtenhancinghumanintent}. To address these limitations, we incorporate force data into human intent estimation, thereby enabling robots to more accurately interpret human intention and enhance their assistance level.

{\noindent \bf Robot Multi-task Learning.} Robot multi-task learning has emerged as a pivotal research domain, evolving from initial grasping tasks to diverse daily applications\cite{zhaoLearnGraspIntention2023, Zhao-RSS-23} and even extending to robot manipulation integrated with language understanding\cite{shridharPerceiveractorMultitaskTransformer2023, bharadhwajRoboAgentGeneralizationEfficiency2024}. Current researches have focused on developing Vision-Language-Action (VLA) models that co-train robot policies with auxiliary tasks to generalize effectively to unseen scenarios\cite{kim2024openvla}. These models typically employ high-capacity transformer architectures incorporating pre-trained foundational models\cite{pmlrv139radford21a, 10377550} to learn multi-task policies. However, their increasing complexity often obscures the role of essential components. Addressing these limitations, Haldar et al.\cite{NEURIPS2024_ff887781} introduced BAKU, a unified architecture that efficiently learns multi-task policies by meticulously integrating observation trunk, action chunking, multi-sensory observations, and action heads, with experimental validation of the synergy among these modules.

Despite these advances, most existing methods still rely on predefined collaboration modes and exhibit constrained capability in seamlessly transitioning based on task requirements and human intentions. Therefore, we extend multi-task policy learning to HRC by deriving collaboration policies from a handful of demonstrations and integrating pHRI with remote cooperation to enable autonomous inference of HRC modes.

\section{Intent-Driven Adaptive Generalized Collaboration}

This section introduces the IDAGC framework, which leverages multimodal information with human intention to facilitate adaptive policy learning by autonomously inferring collaboration modes and dynamically adjusting actions.

\subsection{Problem Statement and Framework Overview}

\begin{figure*}[!t]
    \centering
    \includegraphics[width=0.95\textwidth]{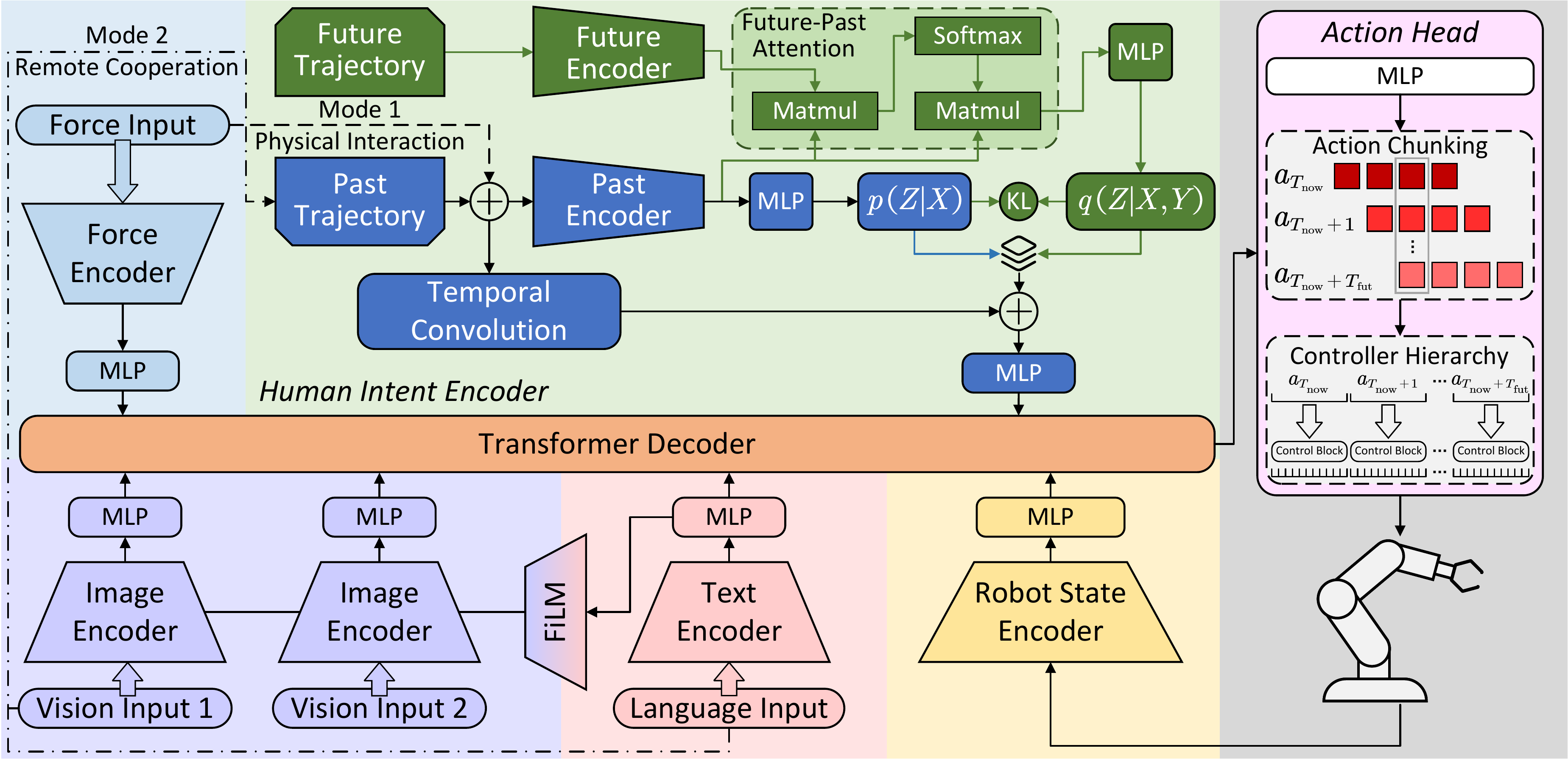}
    \caption{The structure of IDAGC. The green lines denote the train-only operations, while the blue lines denote the predict-only operations. The dashed line denotes pHRI mode, and the dashdotted lines denote remote cooperation mode. A common image encoder is used to process multiple camera views.}
    \label{fig:2.1}
\end{figure*}

This paper investigates HRC through synergistic integration of pHRI and remote cooperation, wherein robots dynamically adapt to task requirements and human intention by seamlessly combining non-contact guidance with physical interaction. To achieve this, we propose IDAGC, a multimodal policy learning framework that predicts human intention, determines appropriate cooperation modes, and optimally adjusts robotic actions, as illustrated in Fig. \ref{fig:2.1}.

IDAGC implements a Transformer-based predictive model to capture interdependencies among vision, language, force, and robot state data, thereby enabling accurate human intent recognition and seamless collaboration mode transitions. In pHRI, the human intent encoder (HIE) utilizes human trajectories and interaction forces to model complex human dynamics and enhance intent prediction accuracy. In remote cooperation, the integration of vision and language input facilitates future trajectory prediction.

To effectively learn multi-task policies that are responsive to human intention and environmental context, our framework incorporates dedicated encoders for each modality, projecting their outputs into a unified latent space. It employs a modified FiLM-conditioned ResNet-18 vision encoder for multi-view camera inputs, a pre-trained text encoder for task instructions, a Multilayer Perceptron (MLP) encoder for robot states, and a causal convolution encoder for force data. 

Additionally, we incorporate a HIE that employs the Conditional Variational Autoencoder (CVAE) framework and introduce latent variable $Z$ to predict the future trajectory of human. We model it as a conditional distribution of the past by learning the multimodal probability distribution function $p(Y|X)$. Thus, the latent distribution can be expressed as
\begin{equation}
    p(Y|X)=\int p(Y|X,Z)p(Z|X)dZ
    \label{2.1}
\end{equation}
where $p(Y|X,Z)$ is the conditional likelihood model and $p(Z|X)$ is the Gaussian conditional prior; $X$ is the past data, where $X=[x^t,\dot{x}^t,f_h^t]_{t=T_{\mathrm{now}}-T_{\mathrm{obs}}}^{T_{\mathrm{now}}}$ during pHRI, and $X=[x^t,\dot{x}^t]_{t=T_{\mathrm{now}}-T_{\mathrm{obs}}}^{T_{\mathrm{now}}}$ during remote cooperation; $Y=[x^t, \dot{x}^t]_{t=T_{\mathrm{now}}}^{T_{\mathrm{now}}+T_{\mathrm{fut}}}$ is the future data, in which $T_{\mathrm{now}}$ is current time, $T_{\mathrm{obs}}$ and $T_{\mathrm{fut}}$ are observable and future time step. 

We introduce loss $L$ consisting of the weighted negative evidence-based lower bound (ELBO) and multi-task behavior cloning loss to optimize the model
\begin{equation}
    \begin{aligned}
        L =&L_{\mathrm{ELBO}} + L_{\mathrm{BC}} \\
        =&\alpha D_{KL}(q(Z|Y,X)||p(Z|X))\\
        &- \beta E_{Z\backsim q(Z|Y,X)}[\log p(Y|Z,X)]\\
        &+ \gamma E_{\{o,a,g\}\backsim X}\left\|a-\pi\left(o|g\right)\right\|^{2}\\
    \end{aligned}
    \label{2.2}
\end{equation}
where $\alpha, \beta, \gamma$ are the weighting factors, $\pi$ is the learned behavior policy. Furthermore, we represent goals $g$ as a text description of the task or a goal image in this work.

Then, we project the representations obtained from all modalities to the same dimensionality through additional MLP layers, to facilitate combining the encoded information. Finally, the Transformer decoder integrates the encoded features from these diverse modalities, while the action head leverages the fused features to generate accurate action and intent predictions. Furthermore, we implement action chunking and controller hierarchy to ensure smooth and natural robot behaviors.

\subsection{Mode 1: Physical Human-Robot Interaction}

When the robot engages in physical interaction with human, force data can facilitate the learning of refined skills and enhance the robot's compliance. In this case, we employ causal convolution for temporal encoding, which helps capturing causal relationships from sequential data such as forces. Furthermore, we transform each dimension of force readings into 2D spectrograms\cite{hou2024adaptive} and input them into a ResNet-18 model \cite{7780459} to encode force/torque data.

Additionally, the robot must accurately infer human intention to provide assistance. We encode the probability distribution by a Transformer-based human intent encoder. Specifically, the past encoder encodes past data into a latent distribution by employing multi-head self-attention combined with standard feedforward and norm layers to capture the past features. Then, it is fed into an MLP to generate $p(Z|X)$.

Similarly, future encoder encodes future data into a latent distribution to capture features. Additionally, we incorporate a future-past attention module to update the features by learning the relationship between past and future data. This is accomplished by integrating the past data features $K$ and $V$ with the future data features $Q$, following the standard feedforward and norm layers. Finally, it is fed into an MLP to generate $q(Z|X,Y)$.

Finally, we obtain the latent variable $Z$ by sampling. During training, $Z$ is obtained from $q(Z|X,Y)$ generated by the future encoder. During testing, it is obtained from $p(Z|X)$ generated by the past encoder.

HIE integrates human dynamics into the network, effectively leveraging past information to predict long-term human intention, providing a foresight basis for proactive adjusting robot behavior. Additionally, the incorporation of Transformer-based CVAE architecture significantly improves prediction accuracy. Compared to state-of-the-art (SOTA) methods\cite{songRobotTrajectronTrajectory2024, franceschiLearningHumanMotion2023, maHumanRobotCollaboration2024}, the proposed HIE exhibits notable advantages, particularly when trajectories are deviated by sudden changes in human intentions.

\subsection{Mode 2: Remote Cooperation}

In remote cooperation, robots operate as independent entities, synchronizing their actions with humans through indirect perception without interactions. By leveraging dedicated encoders for diverse multimodal information within a unified transformer architecture, IDAGC enhances multi-task policy learning, thereby improving compliance and accuracy.

In our framework, we use multiple encoders to process data from various sensors. These encoders transform raw sensor inputs into meaningful feature representations, including visual data, robot states, and various task instructions such as text or target images. To capture these multimodal information, we employ a ResNet-18 \cite{7780459} visual encoder to process scene images and a 6-layer version of MiniLM \cite{NEURIPS2020_3f5ee243} from Sentence Transformers \cite{reimers-gurevych-2019-sentence} to encode textual information. For environments with multiple camera views, we use a common visual encoder across all views. Additionally, we incorporate a FiLM \cite{PerezSVDC18} layer to ensure that image tokens can reliably focus on language instructions, thereby preventing the policy from becoming confused by multiple tasks present in the scene. As for robot states, we use a two-layer MLP encoder to process robot proprioception data.

\subsection{Action Prediction}

Subsequently, we project all encoded results into a common dimensional space by MLP and concatenate them. These features are then fed into a Transformer decoder. This decoder integrates information from various sources, including different camera views, robot state, force, human intentions, and task instructions. Furthermore, we incorporate a learnable token into the decoder, while applying appropriate masks to prevent the model from attending to subsequent data during prediction.

Finally, we input the action feature vectors output by the decoder into the action head to predict the corresponding robot actions. We employ MLP as the action prediction module, which has been validated to perform well on data with limited behavioral diversity \cite{NEURIPS2024_ff887781}. Additionally, considering the temporal dependencies of tasks, we incorporate action chunking \cite{Zhao-RSS-23} with exponential temporal averaging to produce smoother behaviors and mitigate covariate shift, which is commonly encountered in low-data policy learning scenarios. Furthermore, we employ the two-layer controller hierarchy in \cite{10610040}, which is effective for handling contact-rich tasks and reducing oscillations.

\section{Experiments}

In this section, we validate the framework through simulations, real-world experiments and comparative analyses. The data collection, simulations, comparison experiments and applications are detailed in supplementary material.

\subsection{Component Validation and Comparative Analysis}

In this part, we validate the performance of IDAGC with respect to multi-task policy learning and human intent estimation in simulation.

\subsubsection{Multi-Task Policy Learning}

\begin{table}[!t]
    \centering
    \caption{Comparisons of multi-task policy learning performance with existing methods on LIBERO benchmark\cite{NEURIPS2023_8c3c6668}.}
    \begin{tabular}{ccc} \hline
    \multirow{2}{*}{Method} & \multicolumn{2}{c}{Success Rate} \\ \cline{2-3}
                            & LIBERO-90  & LIBERO-10  \\ \hline
    MT-ACT\cite{bharadhwajRoboAgentGeneralizationEfficiency2024} & 54.0\%       & 68.0\%     \\
    BAKU\cite{NEURIPS2024_ff887781}                              & 89.9\%       & 85.0\%    \\
    Ours                                                         & 91.1\%       & 91.0\%     \\ \hline
    \end{tabular}
    \label{tab:4.1}
\end{table}

\begin{figure}[!t] 
    \centering
    \includegraphics[width=0.485\textwidth]{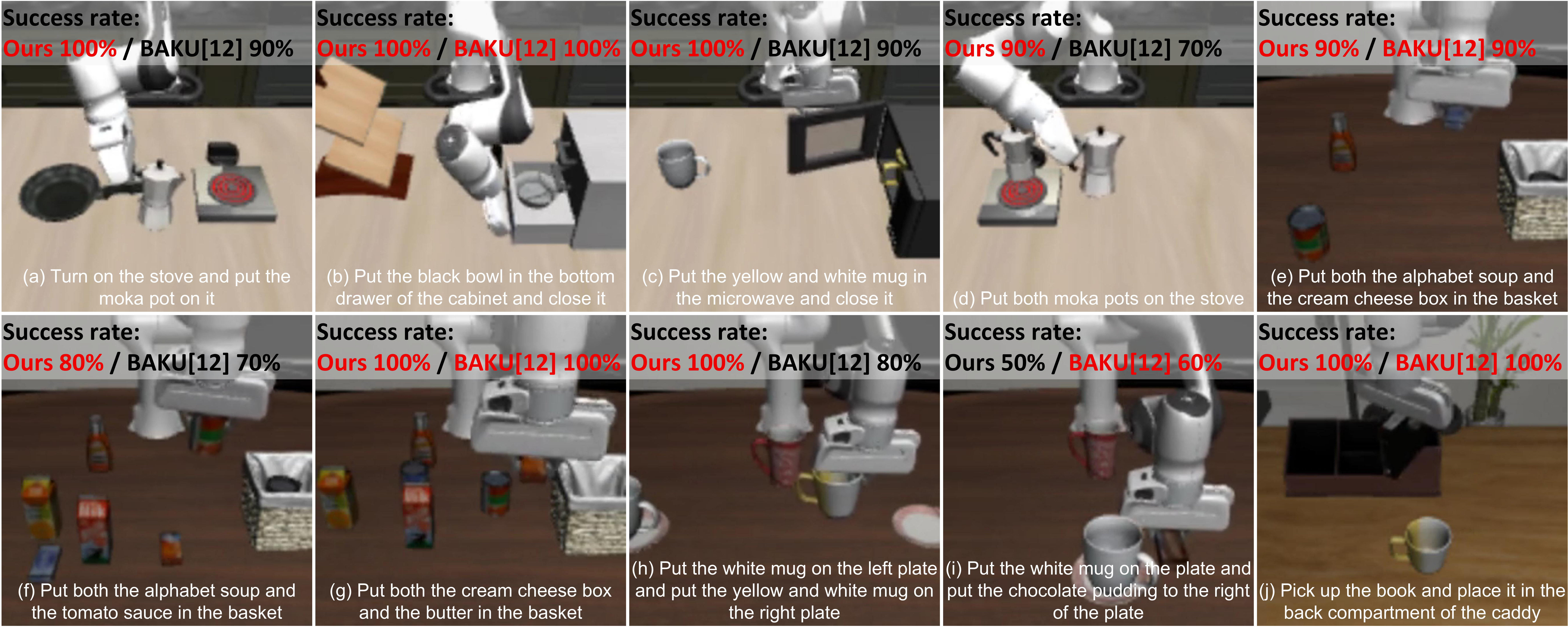}
    \caption{Performance of BAKU\cite{NEURIPS2024_ff887781} and IDAGC on LIBERO-10\cite{NEURIPS2023_8c3c6668}.}
    \label{fig:libero}
\end{figure}



We verify the multi-task policy learning performance of IDAGC in simulation and compare it with SOTA methods, MT-ACT\cite{bharadhwajRoboAgentGeneralizationEfficiency2024} and BAKU\cite{NEURIPS2024_ff887781} on LIBERO benchmark\cite{NEURIPS2023_8c3c6668}. We disable the force encoder and utilize two benchmark manipulation tasks, LIBERO-90 and LIBERO-10, which require the transfer of entangled knowledge. LIBERO-90 comprises 90 manipulation tasks, and LIBERO-10 consists of 10 long-horizon tasks which can test the agent's downstream lifelong learning performance. For each task, we employ 50 demonstrations, incorporating $128\times 128$ images from third-person and gripper camera views, along with robot state as inputs.

As quantified in Table \ref{tab:4.1}, IDAGC achieves SOTA performance. All evaluations are conducted using 10 policies for each task. It is evident that IDAGC outperforms the strongest baseline. On LIBERO-90, IDAGC achieves a 1.2\% improvement. On LIBERO-10, IDAGC improves by 23.0\% and 6.0\% compared to MT-ACT and BAKU. Meanwhile, IDAGC achieves performance comparable to that of short-term manipulation tasks when handling long-horizon tasks. This demonstrates that IDAGC effectively learns planning and execution policies for action sequences.
Furthermore, Fig. \ref{fig:libero} illustrates the tasks included in LIBERO-10 and the success rates of BAKU\cite{NEURIPS2024_ff887781} and the proposed IDAGC on each task. It is obvious that IDAGC demonstrates strong performance than BAKU\cite{NEURIPS2024_ff887781} across most tasks, indicating the effectiveness of the proposed framework in handling diverse task requirements and varying levels of complexity. Overall, IDAGC leverages inter-task relationships, achieving enhanced multi-task learning performance.


\begin{figure}[!t] 
    \centering
    \includegraphics[width=0.485\textwidth]{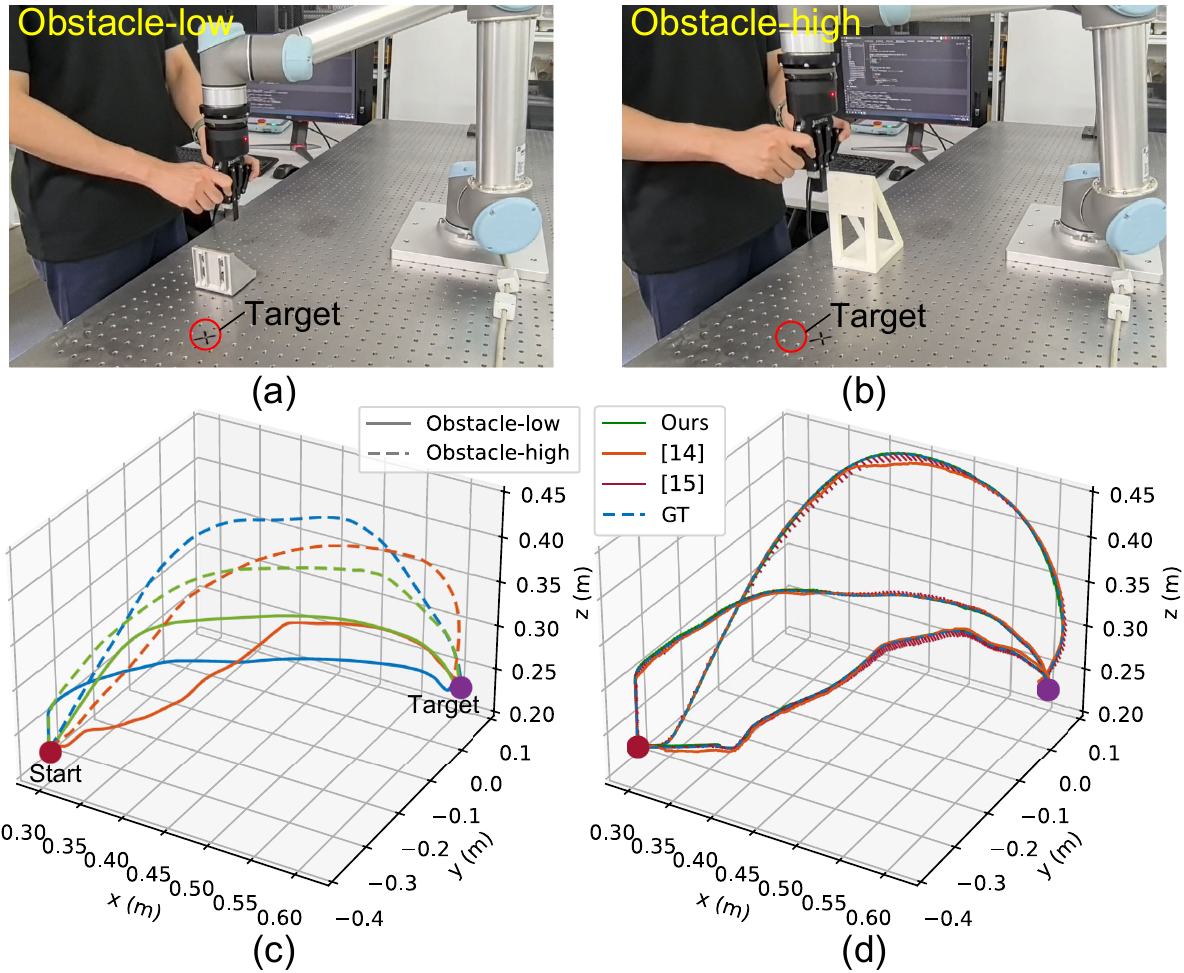}
    \caption{Data collection and predictions. (a)-(c) Data collection processes. (d) Examples of collected trajectories, where solid lines are low obstacle case, and dashed lines are high obstacle case. Each case contains 3 trajectories. (d) Visualization of predictions.}
    \label{fig:data}
\end{figure}

\begin{table}[!t]
    \centering
    \caption{Quantitative comparisons of IDAGC with \cite{cremerModelFreeOnlineNeuroadaptive2020, maHumanRobotCollaboration2024, franceschiLearningHumanMotion2023}.}
    \begin{tabular}{ccccc}
    \hline
    Method & $e_{\mathrm{avg}}\ (\mathrm{mm})$ & $e_{\mathrm{max}}\ (\mathrm{mm})$ \\\hline
    \cite{cremerModelFreeOnlineNeuroadaptive2020}     & 2.56            & 11.03          \\
    \cite{maHumanRobotCollaboration2024}              & 1.94            & 10.45        \\
    \cite{franceschiLearningHumanMotion2023}          & 3.79             & 17.63      \\
    IDAGC                                              & 0.26             & 1.39          \\
    \hline
    \end{tabular}
    \label{tab:4.2}
\end{table}

\subsubsection{Human Intent Estimation}\label{s511}

We validate the effectiveness of IDAGC in real-world pHRI and compare it with \cite{franceschiLearningHumanMotion2023, cremerModelFreeOnlineNeuroadaptive2020, maHumanRobotCollaboration2024} to verify the intent estimation accuracy of our framework. In the experiment, robot and human move along a shared trajectory with an obstacle that robot cannot actively avoid. Upon noticing the obstacle, human guides the robot away from danger, simulating intention changes.

In data collection, we used UR5 in teach mode to gather robot trajectories and forces during human-guided obstacle avoidance, as shown in Fig. \ref{fig:data} (a)-(b). The manipulator is equipped with a Robotiq FT300S Force/Torque Sensor to measure the force. In the scenario, robot and human move along a shared trajectory with an obstacle that the robot cannot actively avoid, and human guides robot away from danger. The end-effector was moved from the start position to the target. We used two randomly set positional obstacles with different heights and collected a total of 50 sets of data at 100 Hz, with 25 sets for each obstacle. The collected trajectories were divided into two datasets, among which 80 were for training, and 20 were for testing in simulations. In simulation, we train the proposed framework and validate the human intent estimation accuracy.

\begin{figure}[!t] 
    \centering
    \includegraphics[width=0.485\textwidth]{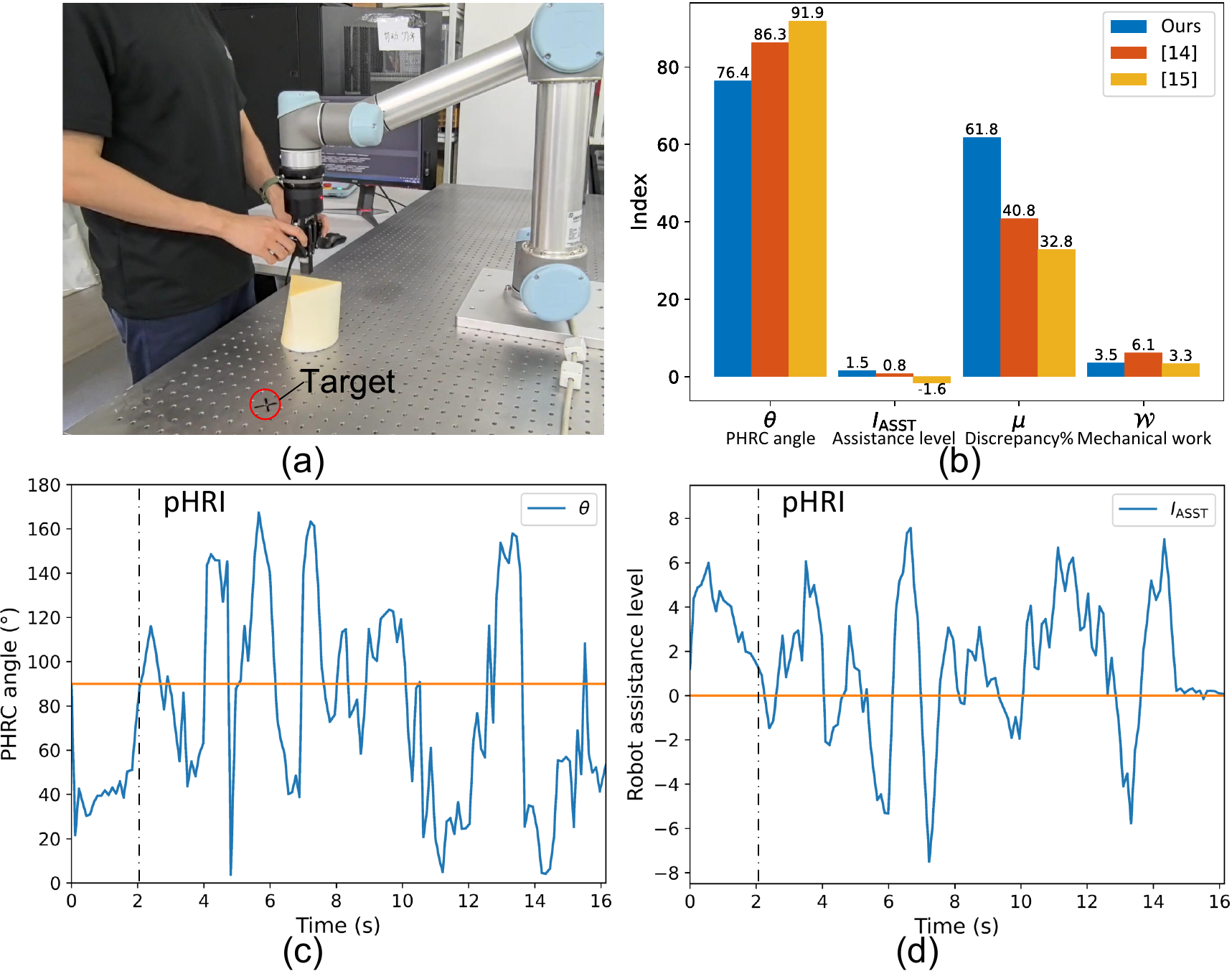}
    \caption{Process and results of pHRI. (a) pHRI process. (b) Comparison of IDAGC with existing methods in terms of metrics. (c) PHRI angle of IDAGC. (d) Robot assistance level of IDAGC.}
    \label{fig:2.5}
\end{figure}

First, we train the proposed framework and validate the human intent estimation accuracy. We compare the prediction accuracy of IDAGC with \cite{franceschiLearningHumanMotion2023, cremerModelFreeOnlineNeuroadaptive2020, maHumanRobotCollaboration2024} in simulation. Since the methods in \cite{cremerModelFreeOnlineNeuroadaptive2020,maHumanRobotCollaboration2024} only support short-term prediction, we use the average prediction error $e_{\mathrm{avg}}$ and max prediction error $e_{\mathrm{max}}$ as evaluation metrics. The results are shown in Table \ref{tab:4.2}. With the increased sampling frequency, the prediction accuracies of all models have improved. Furthermore, Fig. \ref{fig:data} (d) illustrates the prediction results of the above methods on three example trajectories. Clearly, IDAGC excels in trajectory prediction within the pHRI scenario, yielding results that closely match the ground truth, with $e_{\mathrm{avg}}=0.26\ \mathrm{mm}$ and $e_{\mathrm{max}}=1.39\ \mathrm{mm}$, significantly outperforming the existing methods. The incorporation of human intent encoder enables effective modeling of complex and ambiguous intentions, providing latent variables for the subsequent policy module to facilitate collaborative mode switching and policy learning.

\begin{figure*}[!t] 
    \centering
    \includegraphics[width=0.985\textwidth]{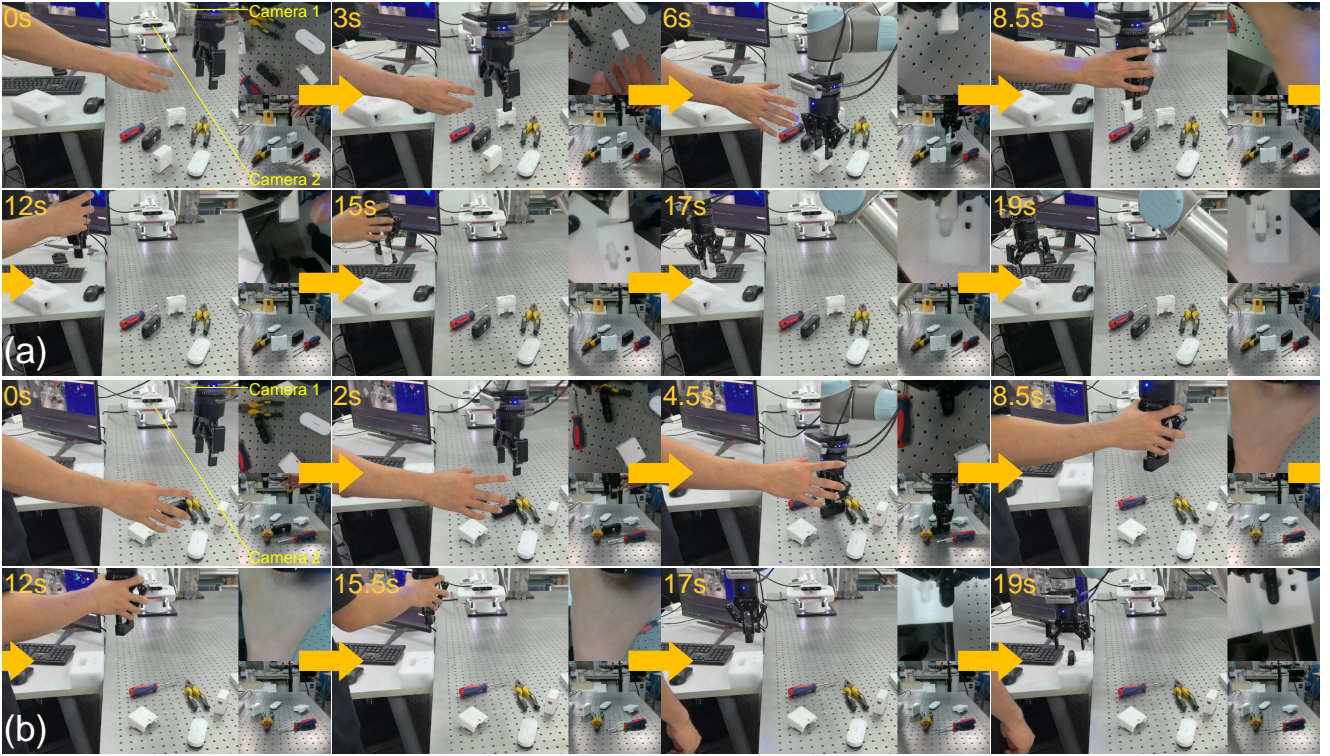}
    \caption{Processes of placing tasks. (a) The robot grasps the white plug ($t=0\sim 6$ s), moves it near the human hand ($t=6\sim 8.5$ s), physically interacts with human ($t=8.5\sim 15$ s), and places the white plug into the slot ($t=15\sim 19$ s). (b) The robot grasps the black camera ($t=0\sim 4.5$ s), moves it near the human hand ($t=4.5\sim 8.5$ s), physically interacts with human ($t=8.5\sim 15.5$ s), and places the black camera into the slot ($t=15.5\sim 19$ s).
    }
    \label{fig:place}
\end{figure*}

\subsection{Physical Human-Robot Interaction}\label{s512}

In this part, to better demonstrate the effectiveness of IDAGC in real-world pHRI, we validate in the same scenarios as Section \ref{s511}. We apply the trained framework and compare with existing methods \cite{franceschiLearningHumanMotion2023, cremerModelFreeOnlineNeuroadaptive2020, maHumanRobotCollaboration2024}. The high-level learning policy controller sends control targets at 10 Hz for the low-level impedance controller at 100 Hz.

Specifically, the evaluation used the following metrics: pHRI angle $\theta=\arccos\frac{f_r^Tf_h}{||f_r|||||f_h||}$, representing the average angle between $f_h$ and $f_r$; robot assistance level index $\mathcal{I} _{\mathrm{ASST}}=\frac{f_{r}^{T}f_{h}}{||f_{h}||}$, denoting the projection of $f_r$ onto $f_h$; human-robot assistance percentage $\mu=\frac{T_{\theta<90^{\circ}}}{T}$, indicating the proportion of time when $\theta < 90^{\circ}$ during motion; and human mechanical work $\mathcal{W}=\int_{0}^{T}f \mathrm{d}x$, a smaller value means less human energy consumption. PHRI angle less than $90^{\circ}$ and robot assistance index greater than 0 indicate that the system is in collaborative state. Additionally, a smaller angle and a larger index represent a higher assistance level.

In the experiment, we replace and randomly position the obstacle. Unlike \cite{maHumanRobotCollaboration2024} applying full-time pHRI, \cite{franceschiLearningHumanMotion2023} and our framework switch human-robot roles when needed. Therefore, except for human mechanical work $\mathcal{W}$, the other metrics of \cite{franceschiLearningHumanMotion2023} and our framework are only calculated during the physical interaction phase. Fig. \ref{fig:2.5} shows the experimental results and compares our framework with \cite{franceschiLearningHumanMotion2023, maHumanRobotCollaboration2024} based on the metrics. The results indicate that our framework outperforms existing methods in pHRI scenarios. The average pHRI angle of \cite{franceschiLearningHumanMotion2023} is $91.9^\circ $, the average robot assistance level index is -1.6, and the human–robot system is in a state of mutual assistance for only 32.8\% of the time, implying that robot hinders human movement. And the average pHRI angle of \cite{maHumanRobotCollaboration2024} is $86.3^\circ $, the average robot assistance level index is 0.8, and the human–robot system is in a state of mutual assistance for 40.8\% of the time. While \cite{maHumanRobotCollaboration2024} demonstrates collaboration potential through its pHRI angle and assistance level, its fully human-guided method overlooks the robot's autonomy, resulting in greater human mechanical work. In comparison, our framework achieves an average pHRI angle of $76.4^\circ $, an average robot assistance level index of 1.5, and the human–robot system is in a state of mutual assistance for 61.8\% of the time. Furthermore, IDAGC effectively balances the robot's autonomy with human intention, resulting in only $3.5\ \mathrm{J}$ human mechanical work. Therefore, the proposed framework demonstrates strong pHRI capabilities, effectively reducing human mechanical work while maintaining a high level of assistance.

\subsection{Real-World Long-Horizon HRC Application} 

In this part, we explore the performance of IDAGC in real-world long-horizon HRC tasks. We used two D435i cameras to capture RGB images as vision inputs. These cameras were mounted at different views, with one mounted on the UR5 end-effector for an overhead perspective and the other on the Franka end-effector for a side view. Language inputs are the task descriptions provided by the operator. Force is measured by the Robotiq FT300S Force/Torque Sensor. For each phase of these tasks, we collected 30 sets of data for training.

We divide each task into three stages and verify the real-world deployment capability of our framework. Initially, the human operator remotely cooperated with the robot using hand and language guidance, enabling the robot to position tools or objects near the human hand. Subsequently, the human and the robot physically interact and move to a target together. Finally, the robot autonomously completed the task. 

In the placing task, the robot was required to grasp an object from a cluttered environment based on human instructions and place it at a specified location. However, the target was outside the field of view. To address this, human guided robot to observe the target, allowing the robot to place the object accurately. The processes are shown in Fig. \ref{fig:place}. It can be seen that IDAGC can flexibly switch collaboration modes, effectively handling the visual blind spots issue caused by hardware limitations with human guidance. 

\begin{figure*}[!t] 
    \centering
    \includegraphics[width=0.985\textwidth]{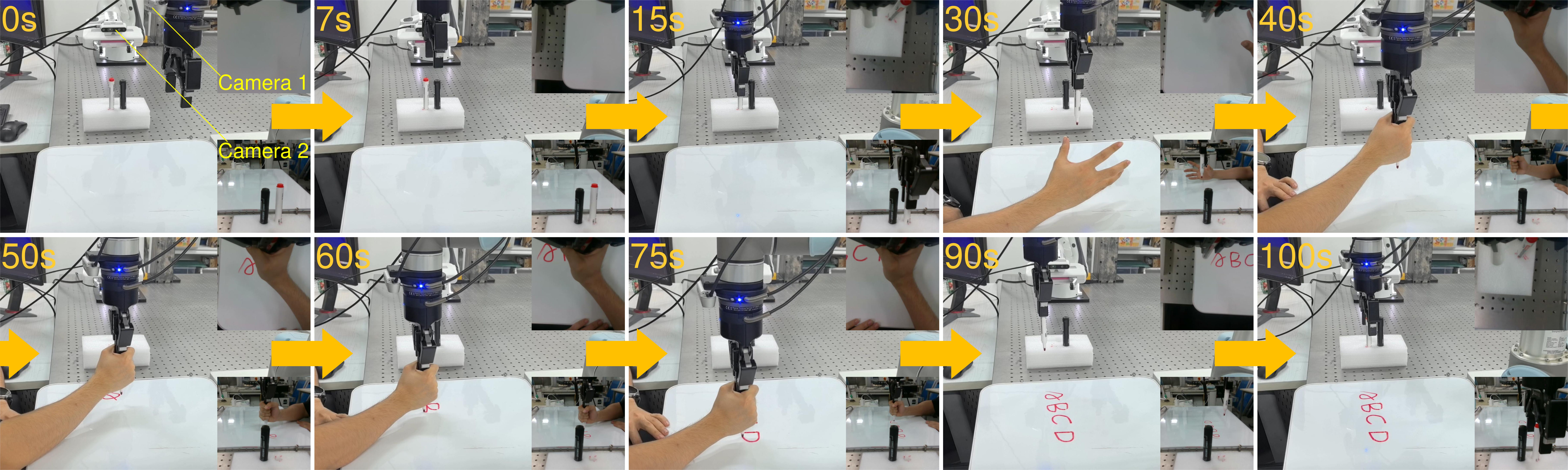}
    \caption{Process of writing task. The robot grasps the red marker ($t=0\sim 15$ s), pulls it out and moves it near the human hand ($t=15\sim 40$ s), physically interacts with human ($t=40\sim 75$ s), and places the red marker into the slot ($t=75\sim 100$ s).
    }
    \label{fig:write}
\end{figure*}

\begin{figure}[!t] 
    \centering
    \includegraphics[width=0.485\textwidth]{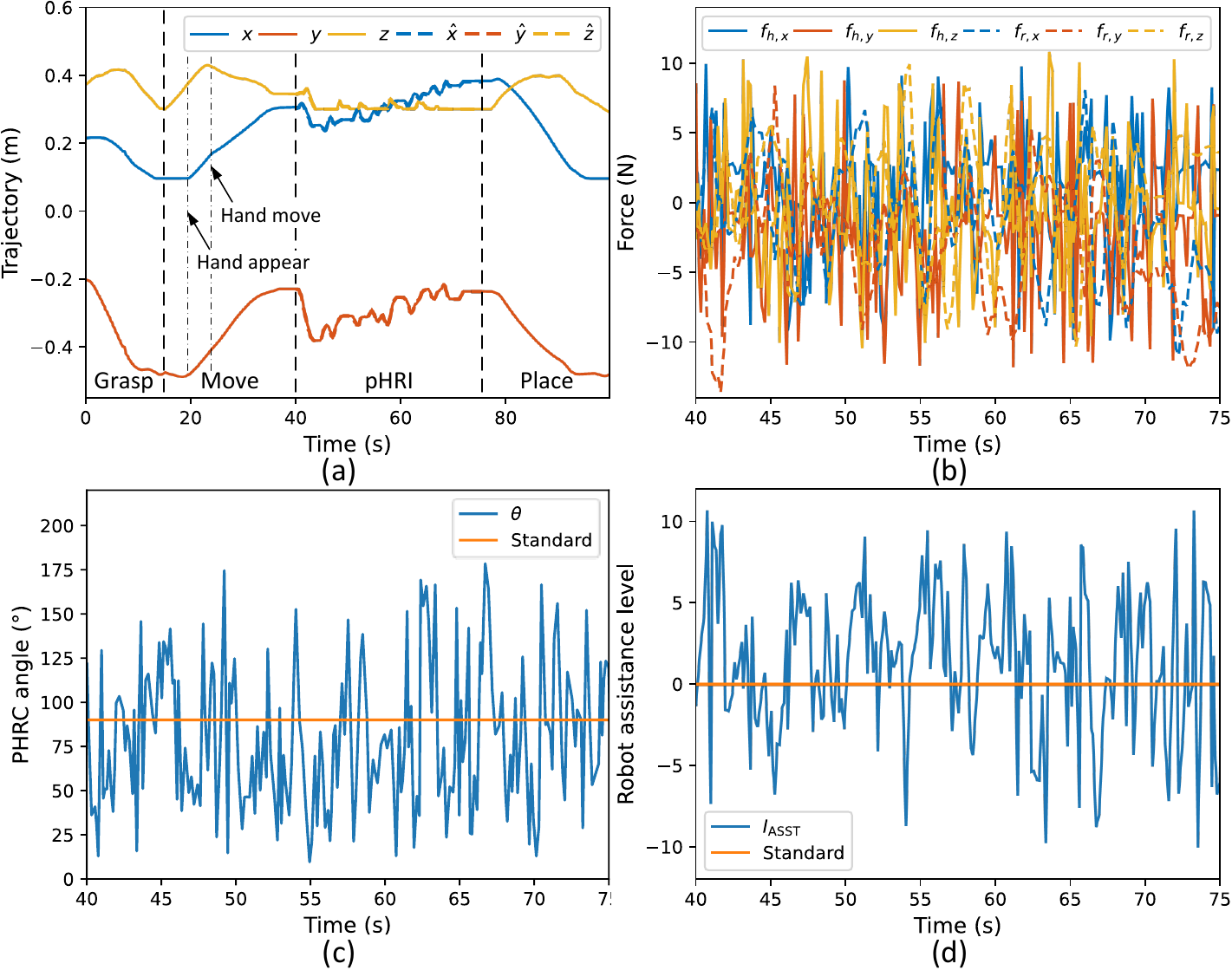}
    \caption{Results of writing task. (a) Predictions of human intention and robot end-effector trajectory of writing task, where the solid line is the robot end-effector trajectory, and dashed line is predict results. (b) Forces applied by human and robot. (c) PHRI angle. (d) Robot assistance level.}
    \label{fig:write_tr}
\end{figure}

Next, in the writing task, the robot first handed the red marker to human and then assisted in writing letters. During writing, the robot estimated human intentions in real-time and adjusted its behavior to minimize human effort, reducing human-robot disagreement and improving assistance level. Once the writing was complete, the human used language to guide the robot to return the marker to its original position. The process of the writing task is shown in Fig. \ref{fig:write}, while the corresponding results are shown in Fig. \ref{fig:write_tr}. The findings indicate that IDAGC predicts accurately, enabling robot to understand human intentions. Notably, as illustrated in Fig. \ref{fig:write_tr} (a), after grasping the marker, the robot transitioned from an upward lifting motion to a handover motion upon detecting the human hand. As the human made slight movements, the robot synchronously adjusted its motion to facilitate the handover. Furthermore, we evaluate the pHRI performance during writing, as shown in Fig. \ref{fig:write_tr} (b)-(d), the average pHRC angle $\theta$ is $78.52^\circ $, the average robot assistance level index $\mathcal{I} _{\mathrm{ASST}}$ is $1.34$. Additionally, the system is in a state of mutual assistance for 64.71\% of the time, highlighting its effectiveness in collaborative scenarios. 


In summary, IDAGC demonstrates strong adaptability to dynamic collaboration requirements by accurately inferring human intentions from multimodal data and flexibly switching between HRC modes in diverse real-world scenarios.

\section{Conclusion}

This paper presents the IDAGC framework, which integrates multimodal data and intent estimation through dedicated encoders. IDAGC autonomously infers collaboration modes and dynamically adjusts robotic actions, effectively overcoming the limitations of existing HRC methods. It accommodates both pHRI and remote cooperation while demonstrating robust multi-task learning from a handful of demonstrations. Simulation and real-world experiments validate that IDAGC significantly enhances intent estimation accuracy and improves the overall comprehensiveness of HRC. Overall, IDAGC holds substantial potential for advancing multimodal integration and adaptive collaboration policies, thereby enhancing the deployment of robotic systems in complex HRC scenarios.

\bibliographystyle{IEEEtran}
\bibliography{IEEEabrv,reference}

\end{document}